\documentclass[runningheads]{llncs}
\usepackage[T1]{fontenc}
\usepackage{graphicx}
\usepackage{comment}
\usepackage{amssymb}
\usepackage{amsmath}
\usepackage{multirow}
\usepackage{booktabs}
\usepackage{color}

\def\overview{
\begin{figure}[t]
\centering
\includegraphics[width=0.8\linewidth]{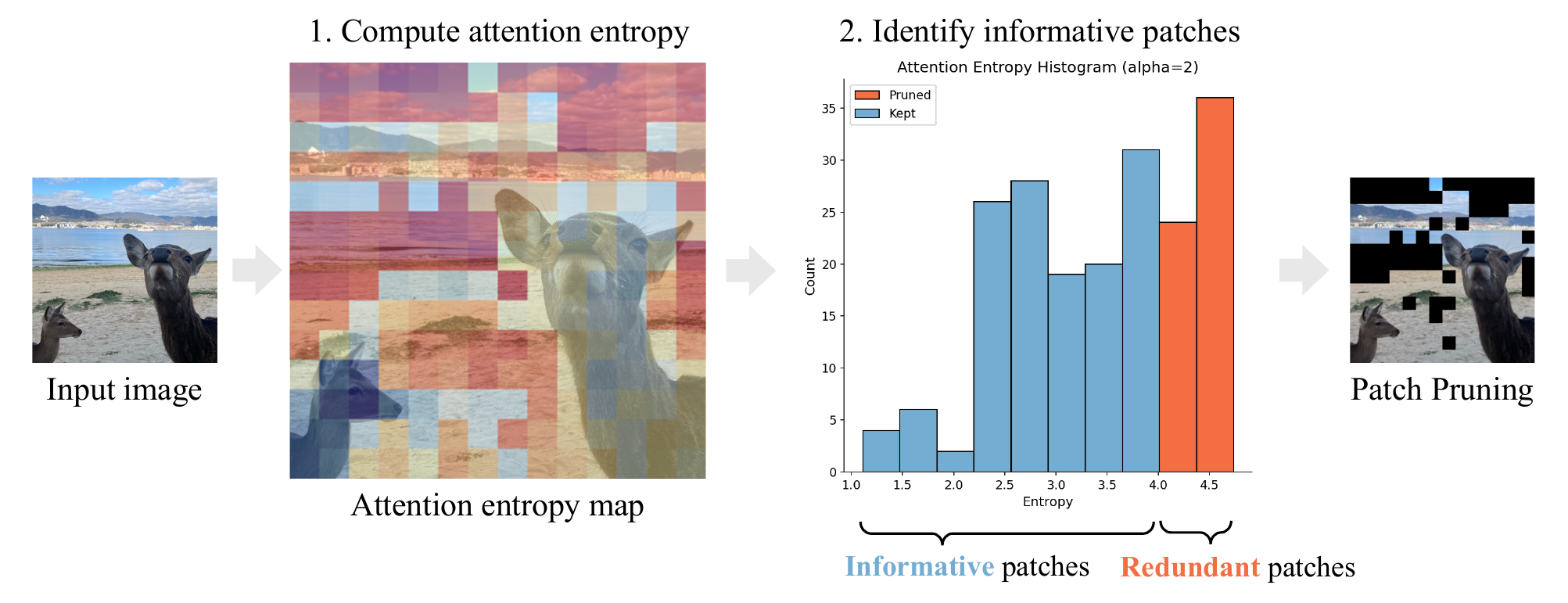}
\caption{Overview of our key idea. The left figure shows an attention entropy map, where red indicates higher entropy and blue indicates lower entropy. We observe that low attention entropy corresponds to foreground regions and high attention entropy to background, which helps identify informative patches. Based on this, we use attention entropy as the pruning criterion, as illustrated on the right.}
\label{fig:overview}
\end{figure}
}

\def\visattnent{
\begin{figure}[t!]
\centering
\includegraphics[width=0.9\linewidth]{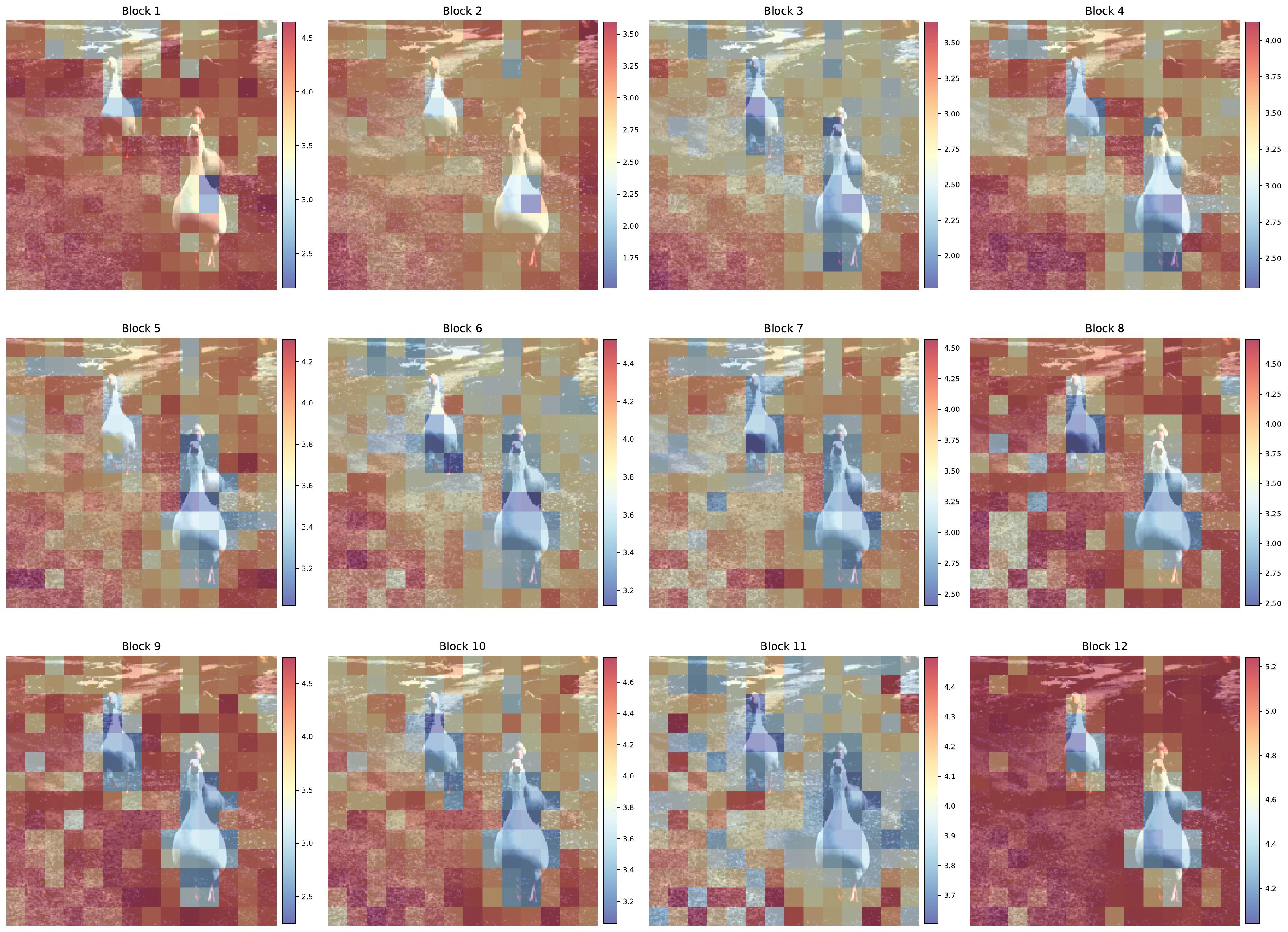}
\caption{Visualization of R\'enyi attention entropy ($\alpha=2.0$) for each Transformer block in DeiT-S. This visualization shows that attention entropy depends on Transformer layer depth, and lower entropy corresponds to foreground regions.}
\label{fig:visattnent}
\end{figure}
}

\def\proposed{
\begin{figure*}[t!]
\centering
\includegraphics[width=0.7\linewidth]{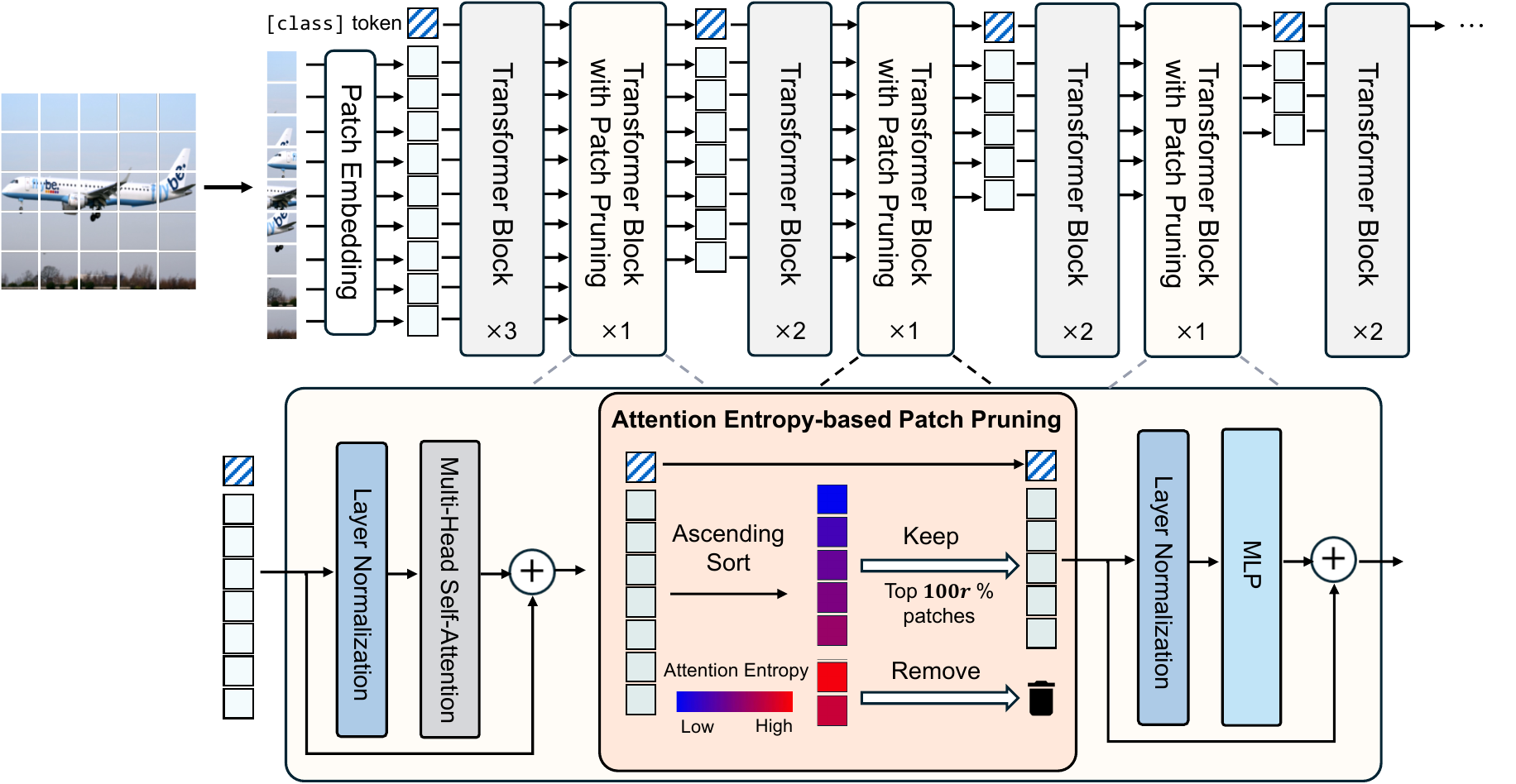}
\caption{Overall pipeline of the R\'enyi attention entropy pruning. The pruning procedure is described in Section~\ref{ssec:patch_pruning}.}
\label{fig:proposed}
\end{figure*}
}

\def\vispatch{
\begin{figure*}[t!]
\centering
  \begin{minipage}[t]{0.5\textwidth}\centering
    \includegraphics[width=\linewidth,height=3.2cm,keepaspectratio]{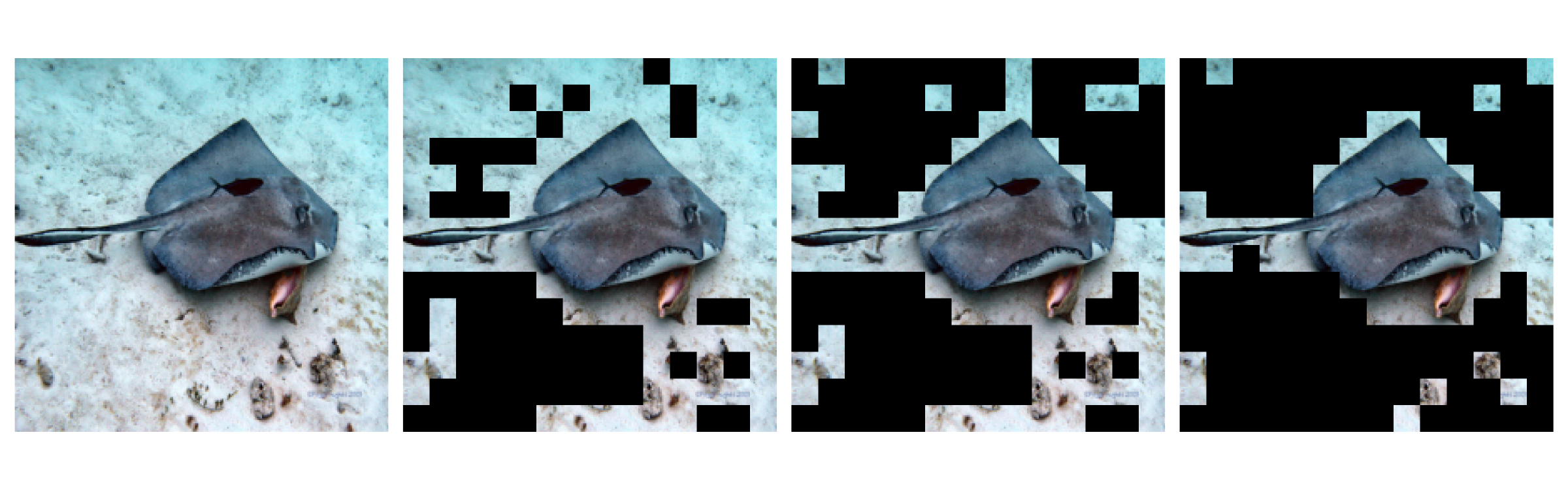}
  \end{minipage}\hfill
  \begin{minipage}[t]{0.5\textwidth}\centering
    \includegraphics[width=\linewidth,height=3.2cm,keepaspectratio]{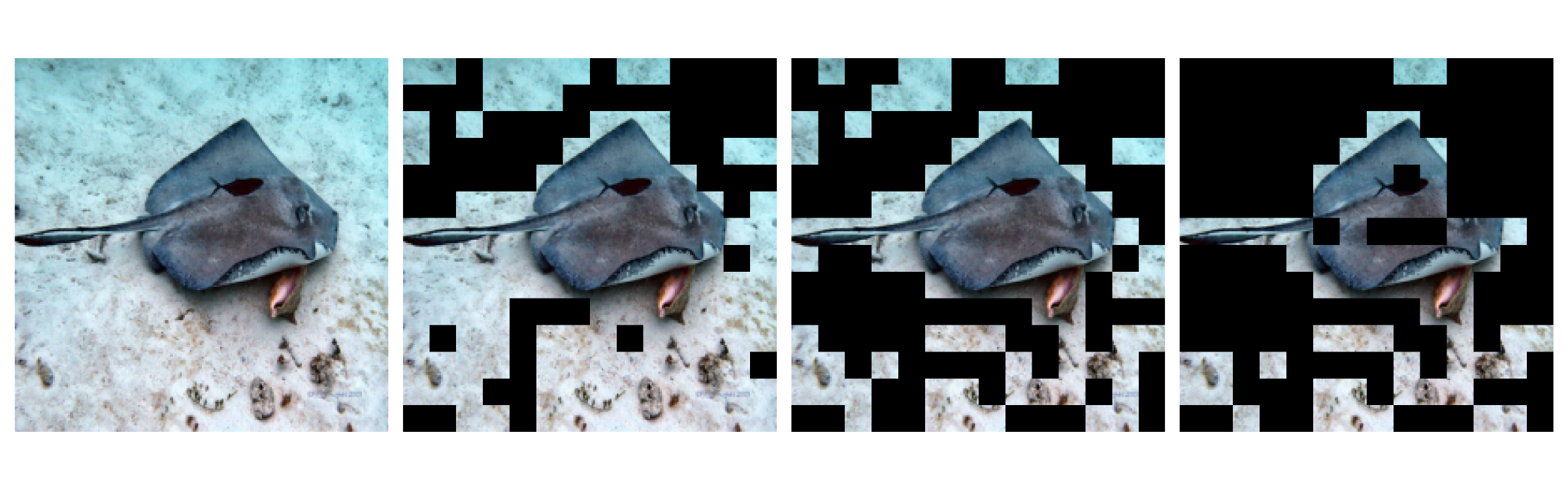}
  \end{minipage}\hfill

  \begin{minipage}[t]{0.5\textwidth}\centering
    \includegraphics[width=\linewidth,height=3.2cm,keepaspectratio]{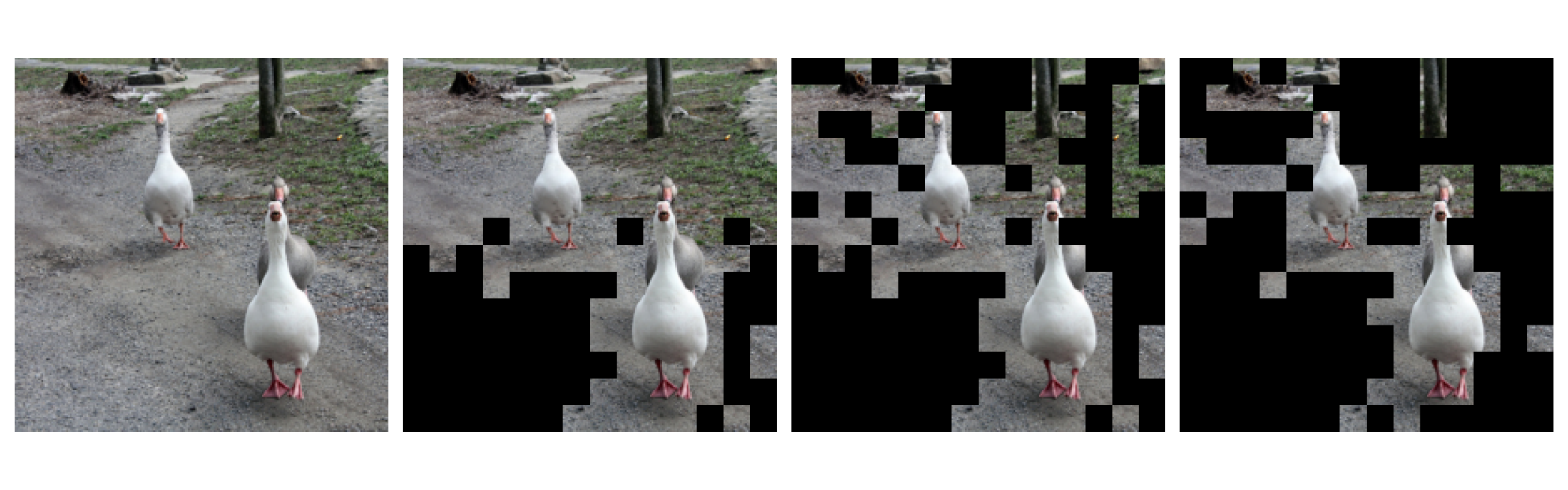}
  \end{minipage}\hfill
  \begin{minipage}[t]{0.5\textwidth}\centering
    \includegraphics[width=\linewidth,height=3.2cm,keepaspectratio]{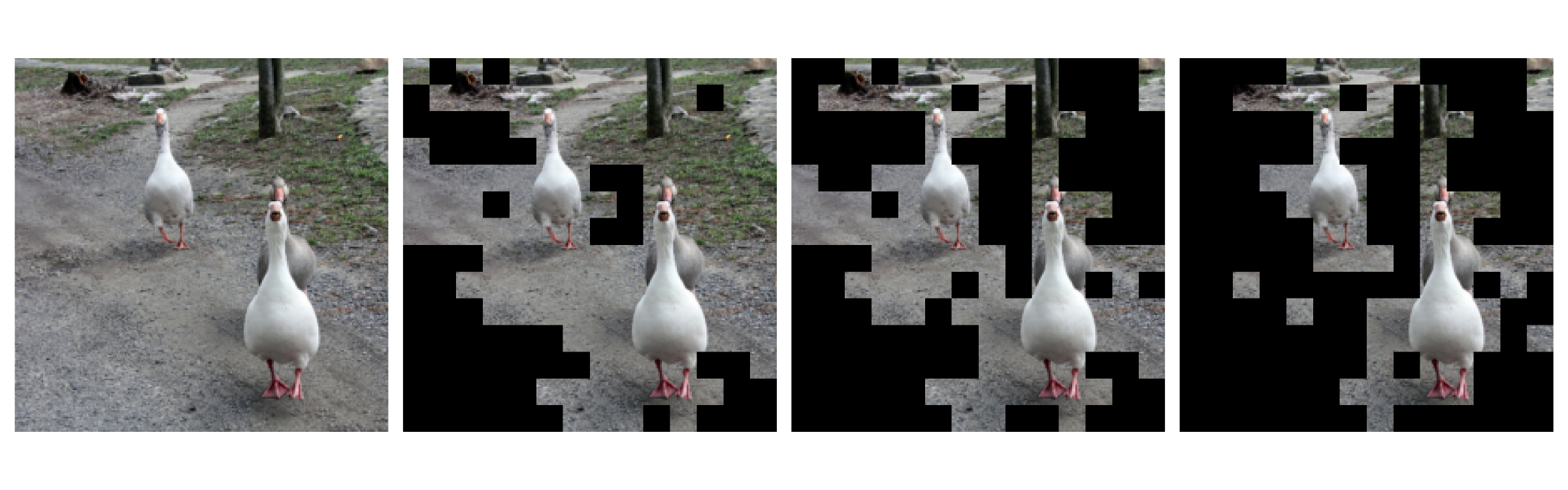}
  \end{minipage}\hfill

    \begin{minipage}[t]{0.5\textwidth}\centering
    \includegraphics[width=\linewidth,height=3.2cm,keepaspectratio]{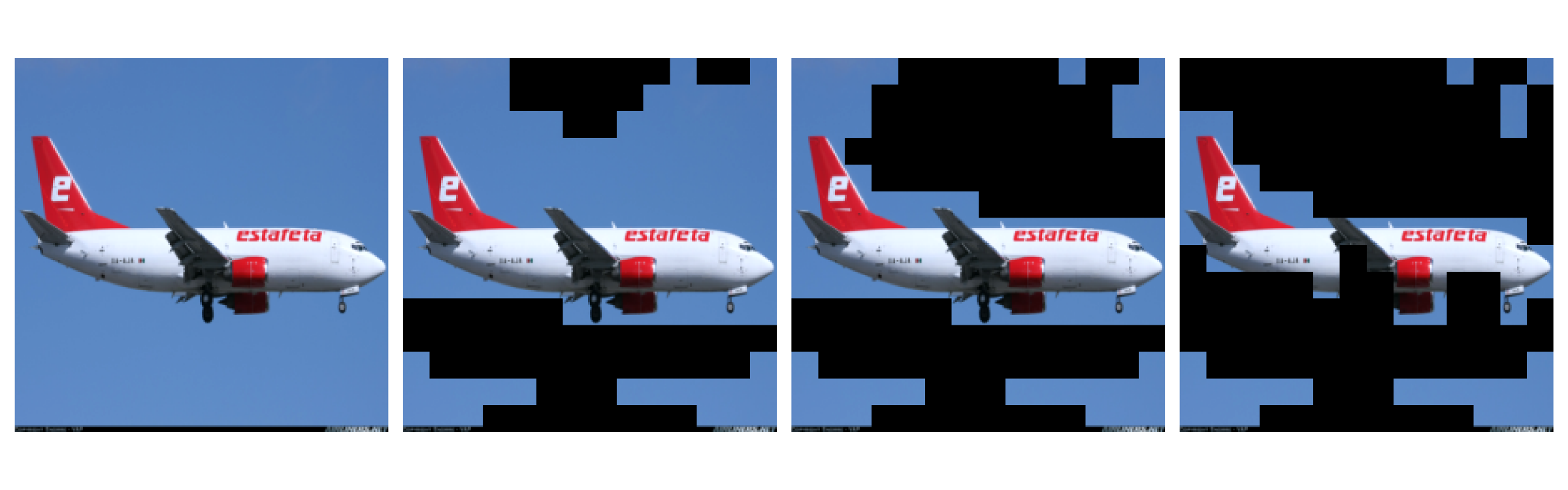}
  \end{minipage}\hfill
  \begin{minipage}[t]{0.5\textwidth}\centering
    \includegraphics[width=\linewidth,height=3.2cm,keepaspectratio]{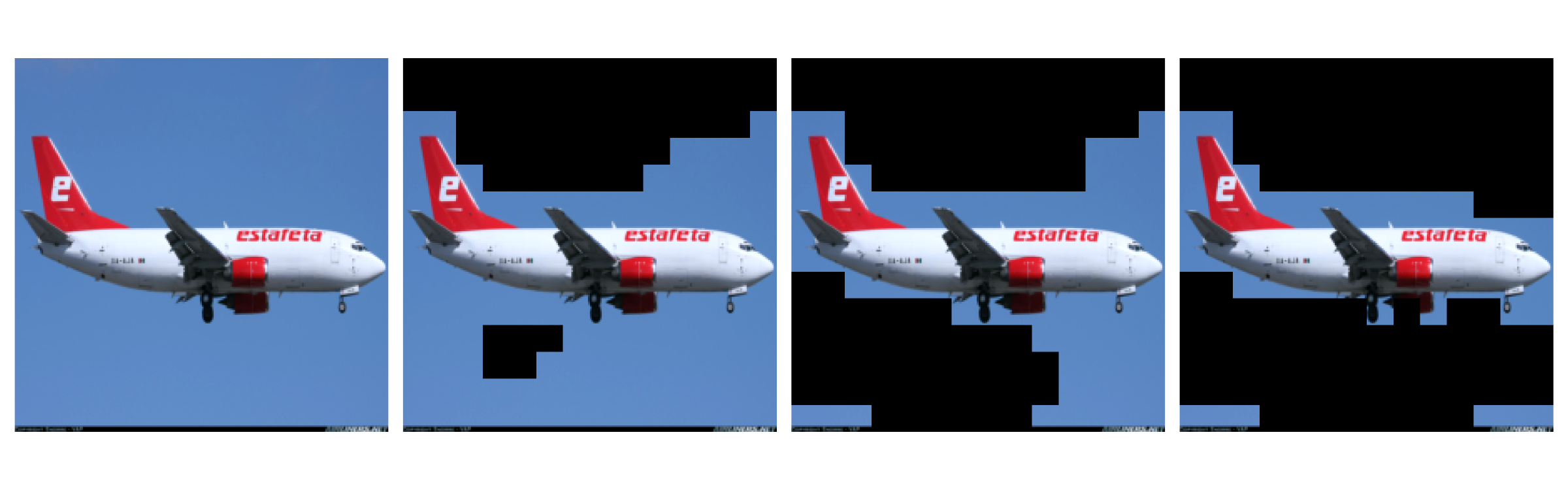}
  \end{minipage}\hfill

    \begin{minipage}[t]{0.5\textwidth}\centering
    \includegraphics[width=\linewidth,height=3.2cm,keepaspectratio]{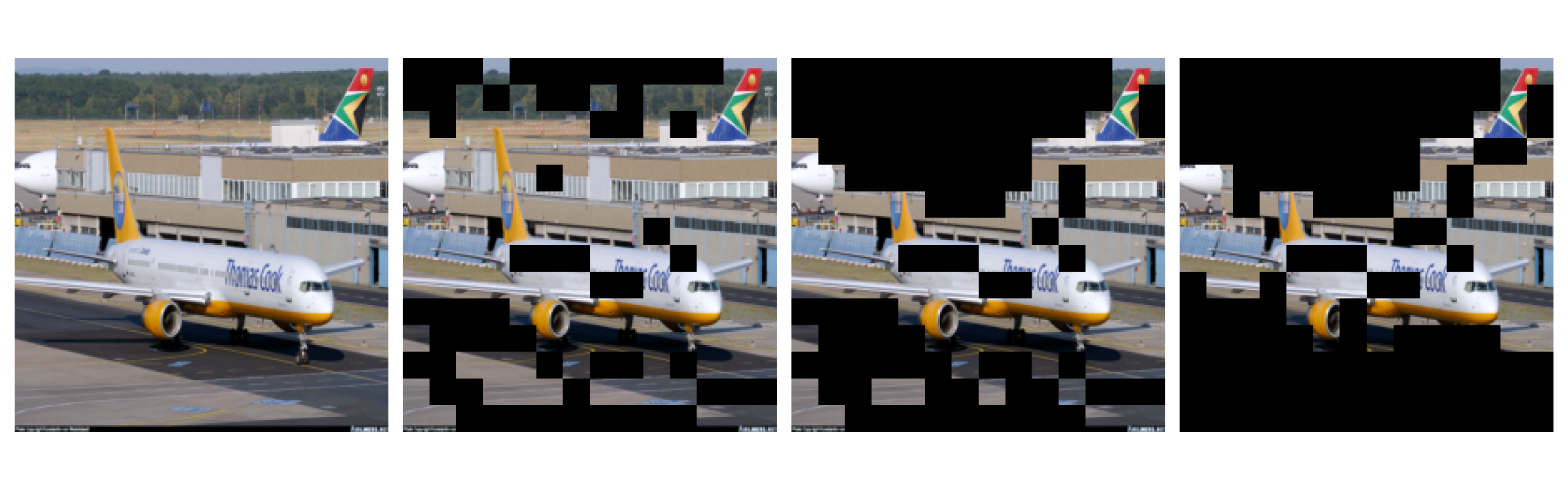}
  \end{minipage}\hfill
  \begin{minipage}[t]{0.5\textwidth}\centering
    \includegraphics[width=\linewidth,height=3.2cm,keepaspectratio]{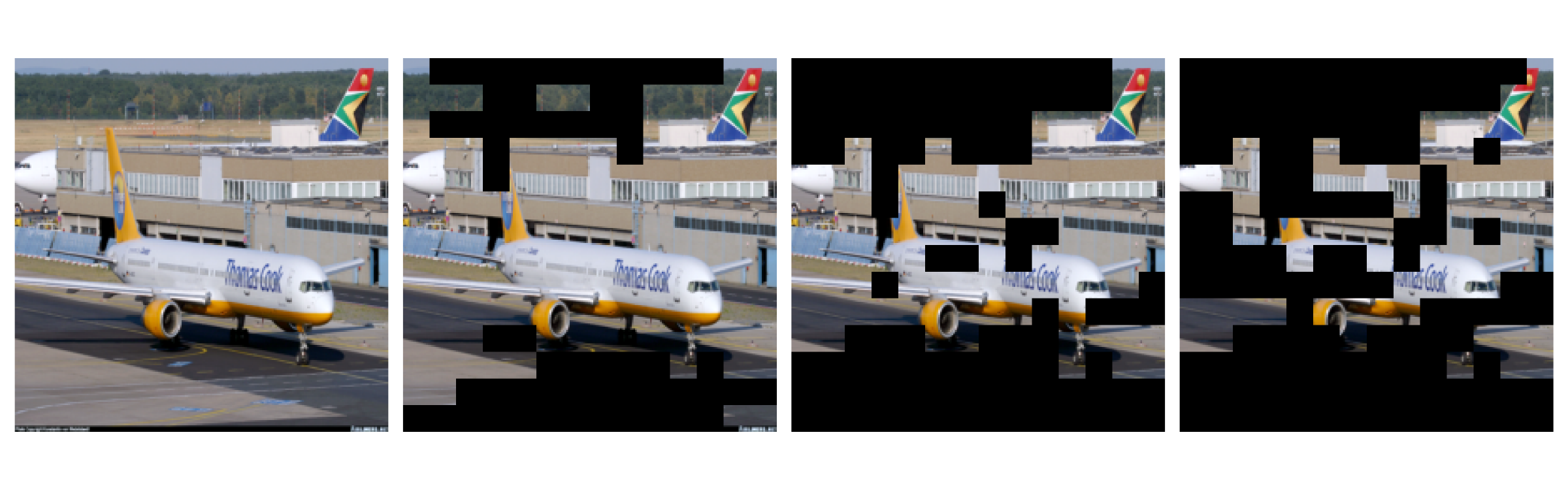}
  \end{minipage}\hfill

    \begin{minipage}[t]{0.5\textwidth}\centering
    \includegraphics[width=\linewidth,height=3.2cm,keepaspectratio]{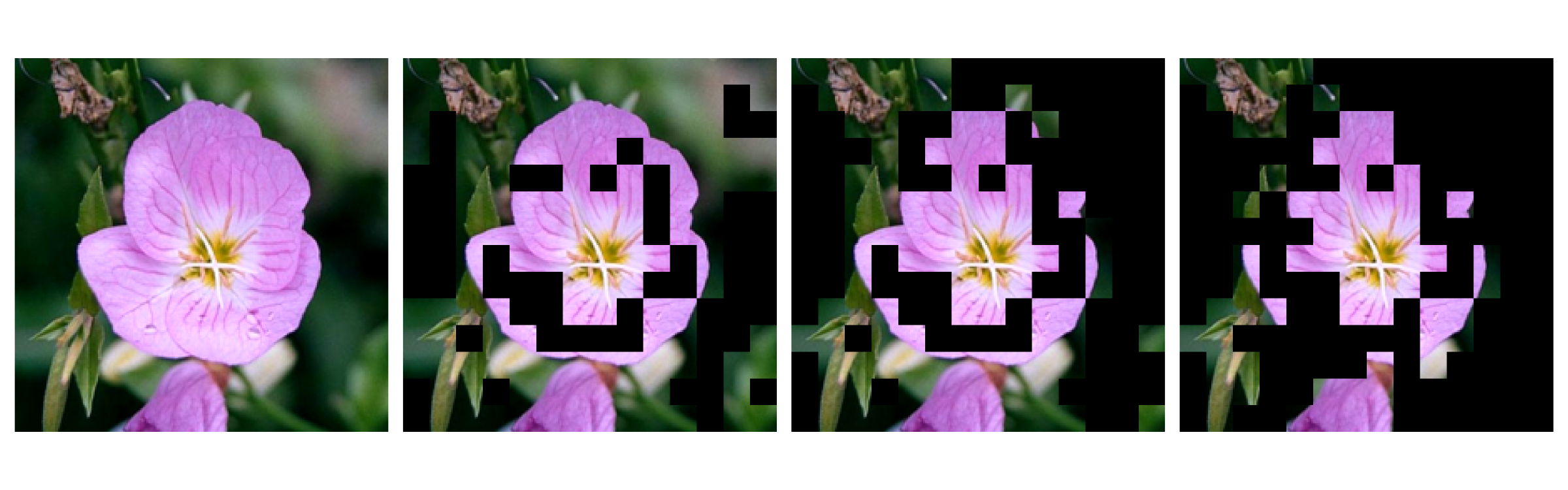}
  \end{minipage}\hfill
  \begin{minipage}[t]{0.5\textwidth}\centering
    \includegraphics[width=\linewidth,height=3.2cm,keepaspectratio]{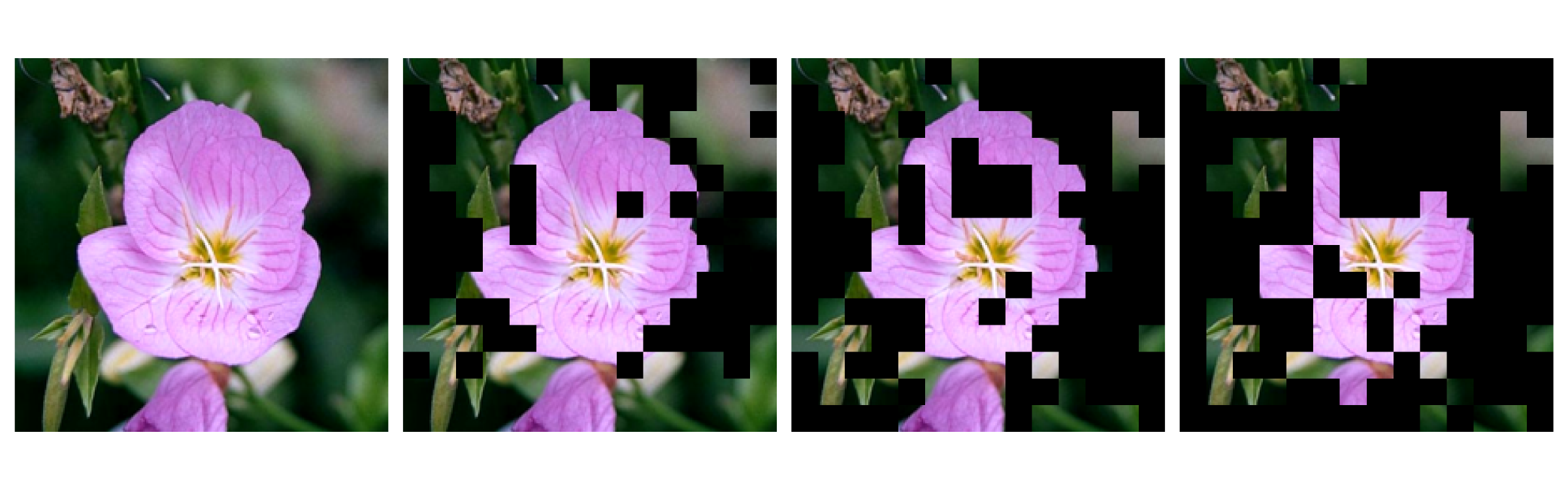}
  \end{minipage}\hfill

    \begin{minipage}[t]{0.5\textwidth}\centering
    \includegraphics[width=\linewidth,height=3.2cm,keepaspectratio]{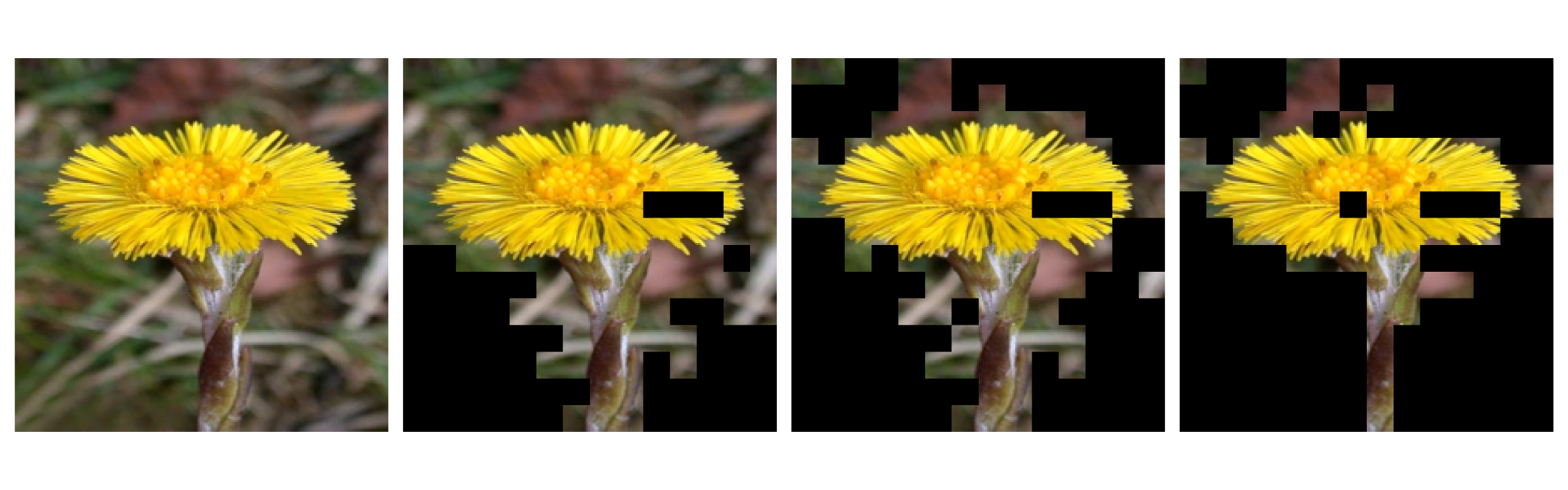}
    \vspace{2pt}\small EViT
  \end{minipage}\hfill
  \begin{minipage}[t]{0.5\textwidth}\centering
    \includegraphics[width=\linewidth,height=3.2cm,keepaspectratio]{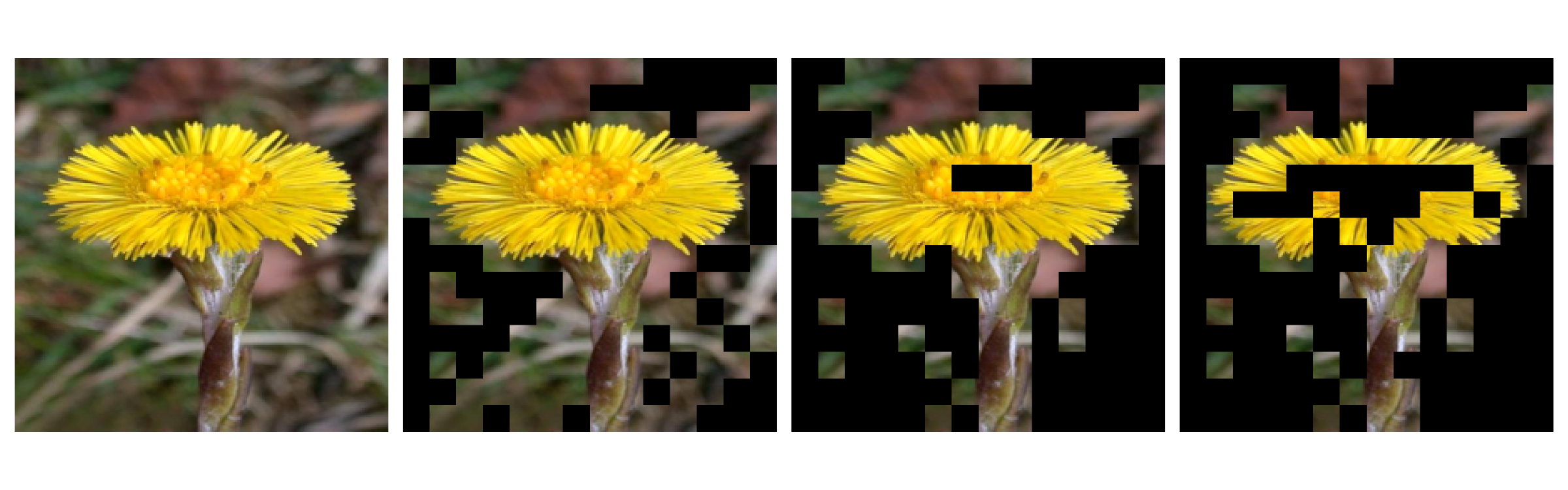}
    \vspace{2pt}\small Ours: R\'enyi Attention Entropy Pruning
  \end{minipage}\hfill
\caption{Visualizations of patch pruning results for EViT and R\'enyi attention entropy-based approach on \texttt{ImageNet-100}, \texttt{FGVC Aircraft}, and \texttt{Oxford Flowers102}. From left to right: input image, pruning results at Blocks 4, 7, and 10. The keep rate is $r=0.7$, and for our method we show the results with the tuned $\alpha$.}
\label{fig:vispatch}
\end{figure*}
}

\def\attnhist{
\begin{figure*}[t!]
\centering
\centering
  \begin{minipage}[t]{0.24\linewidth}
    \centering
    \includegraphics[width=\linewidth]{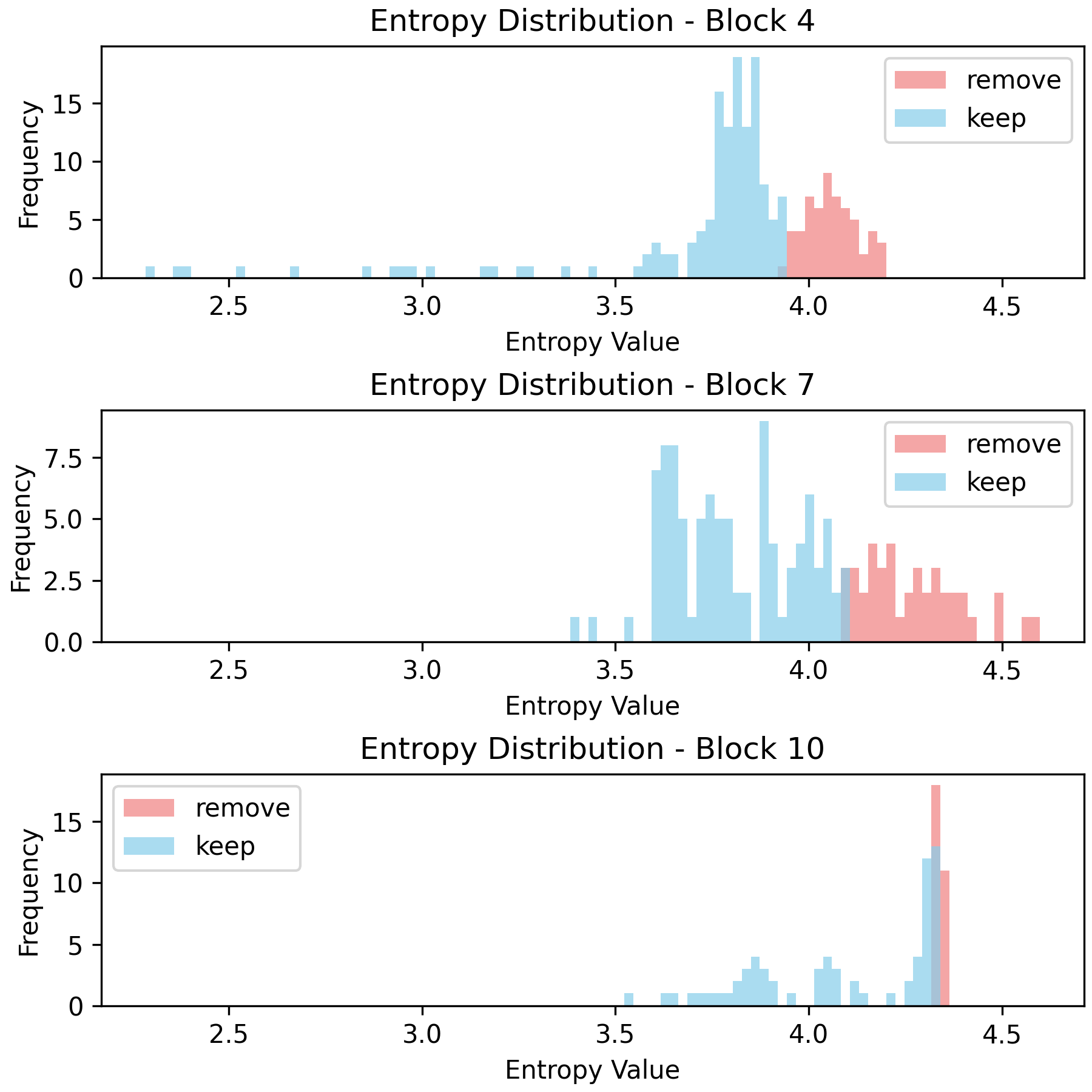}\\
    \small Shannon
  \end{minipage}\hfill
  \begin{minipage}[t]{0.24\linewidth}
    \centering
    \includegraphics[width=\linewidth]{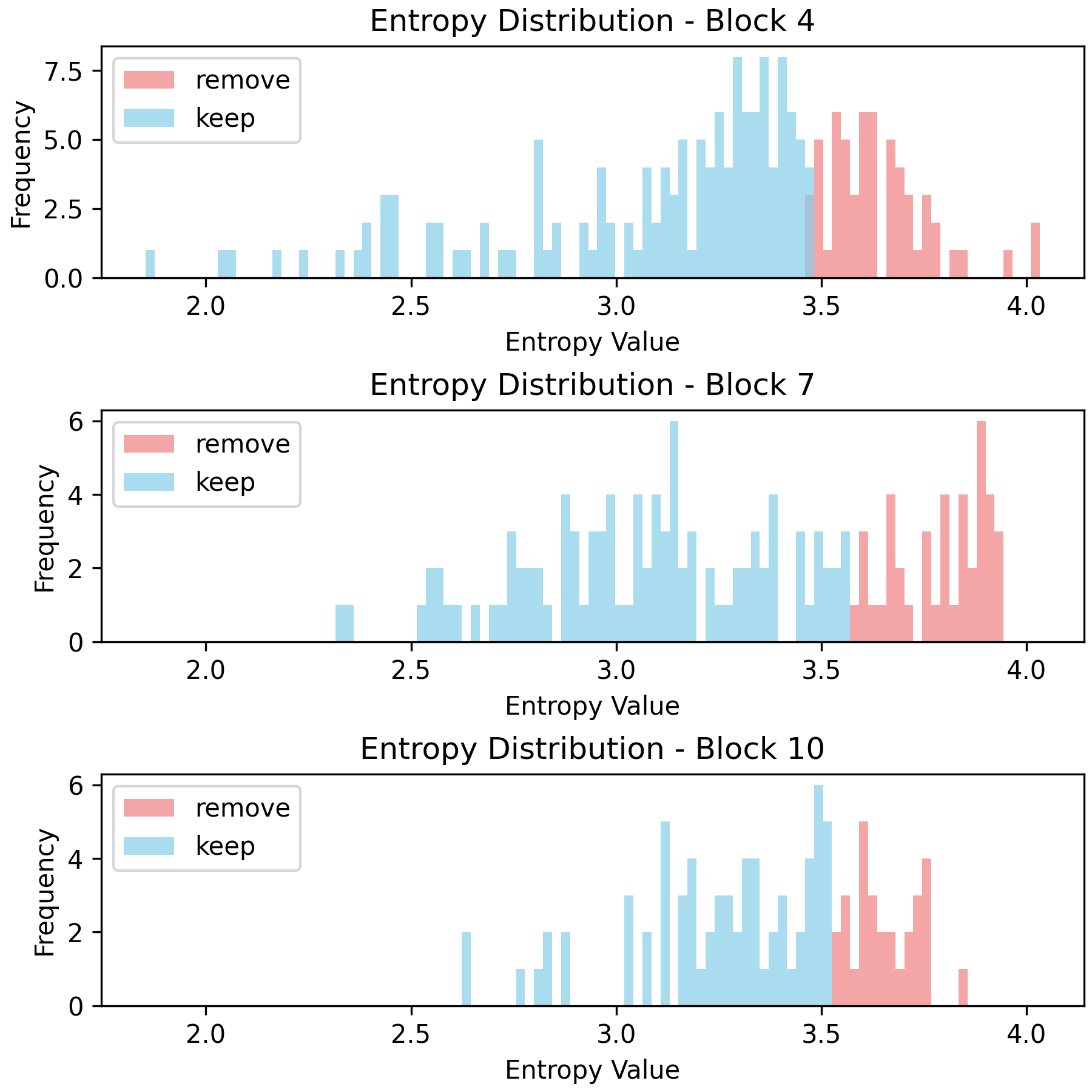}\\
    \small R\'enyi ($\alpha=2.0$)
  \end{minipage}\hfill
  \begin{minipage}[t]{0.24\linewidth}
    \centering
    \includegraphics[width=\linewidth]{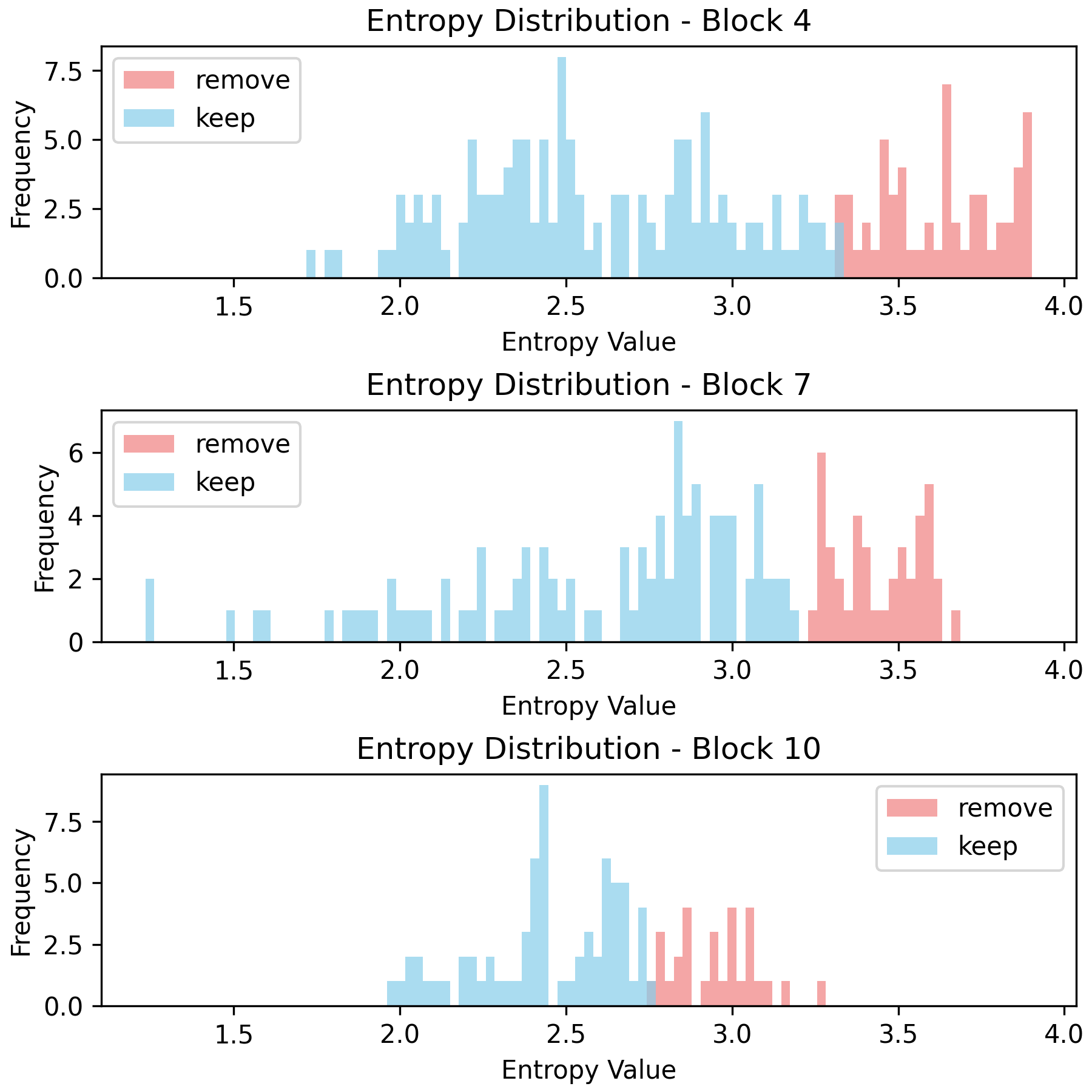}\\
    \small R\'enyi ($\alpha=5.0$)
  \end{minipage}\hfill
  \begin{minipage}[t]{0.24\linewidth}
    \centering
    \includegraphics[width=\linewidth]{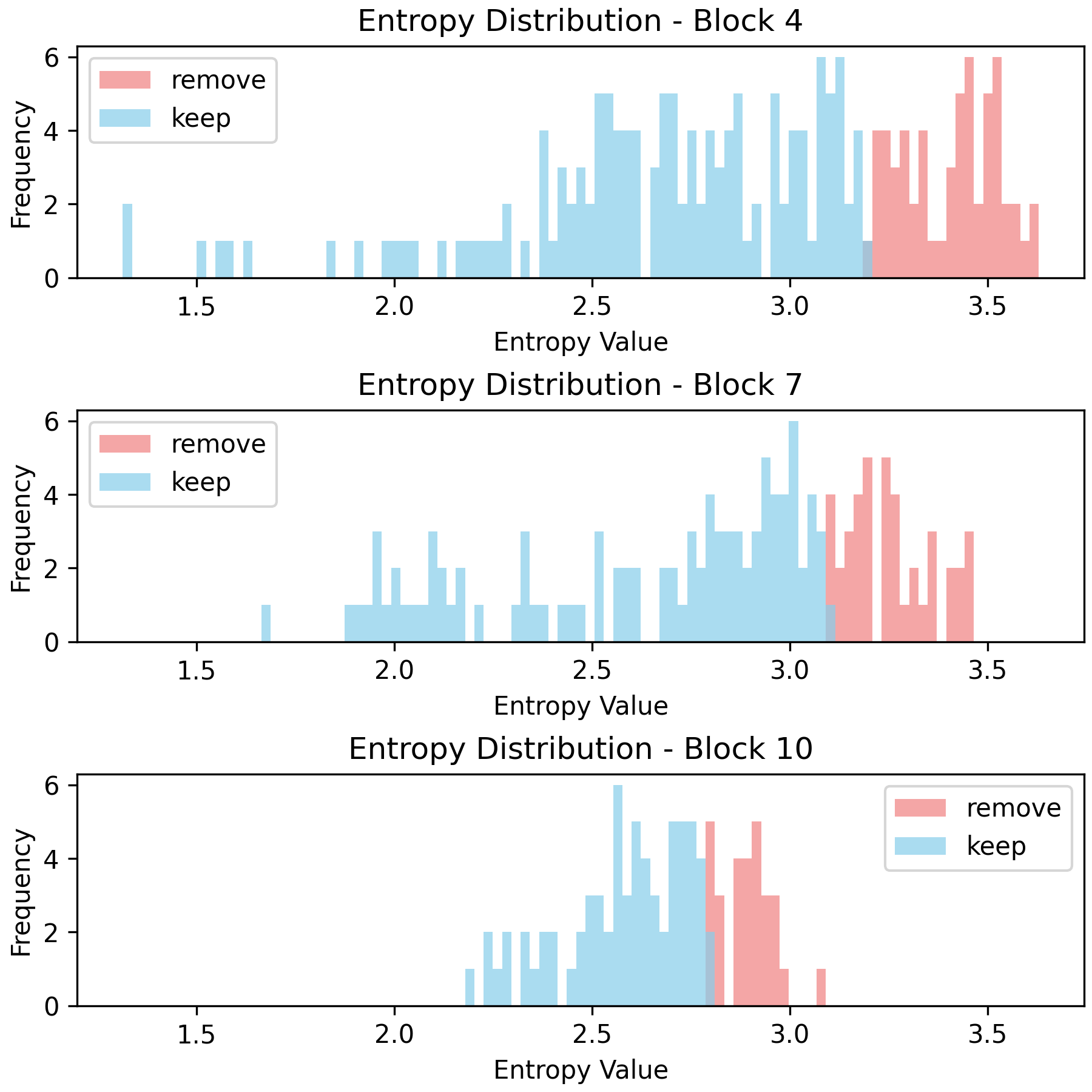}\\
    \small R\'enyi ($\alpha=10.0$)
  \end{minipage}
\caption{Visualization of Shannon and Rényi attention entropies. For each DeiT-S block, the figure shows histograms of Shannon attention entropy ($\alpha=1.0$) and Rényi attention entropy at different $\alpha$ orders. Blue indicates informative patches that are kept, and red indicates redundant patches that are pruned. The results show that the Rényi order controls peak emphasis and allows the characterization of the attention distribution to be adapted to the task.}
\label{fig:attnhist}
\end{figure*}
}

\def\attndist{
\begin{figure}[t!]
  \centering
  \begin{minipage}{0.5\linewidth}
    \centering
    \includegraphics[width=\linewidth]{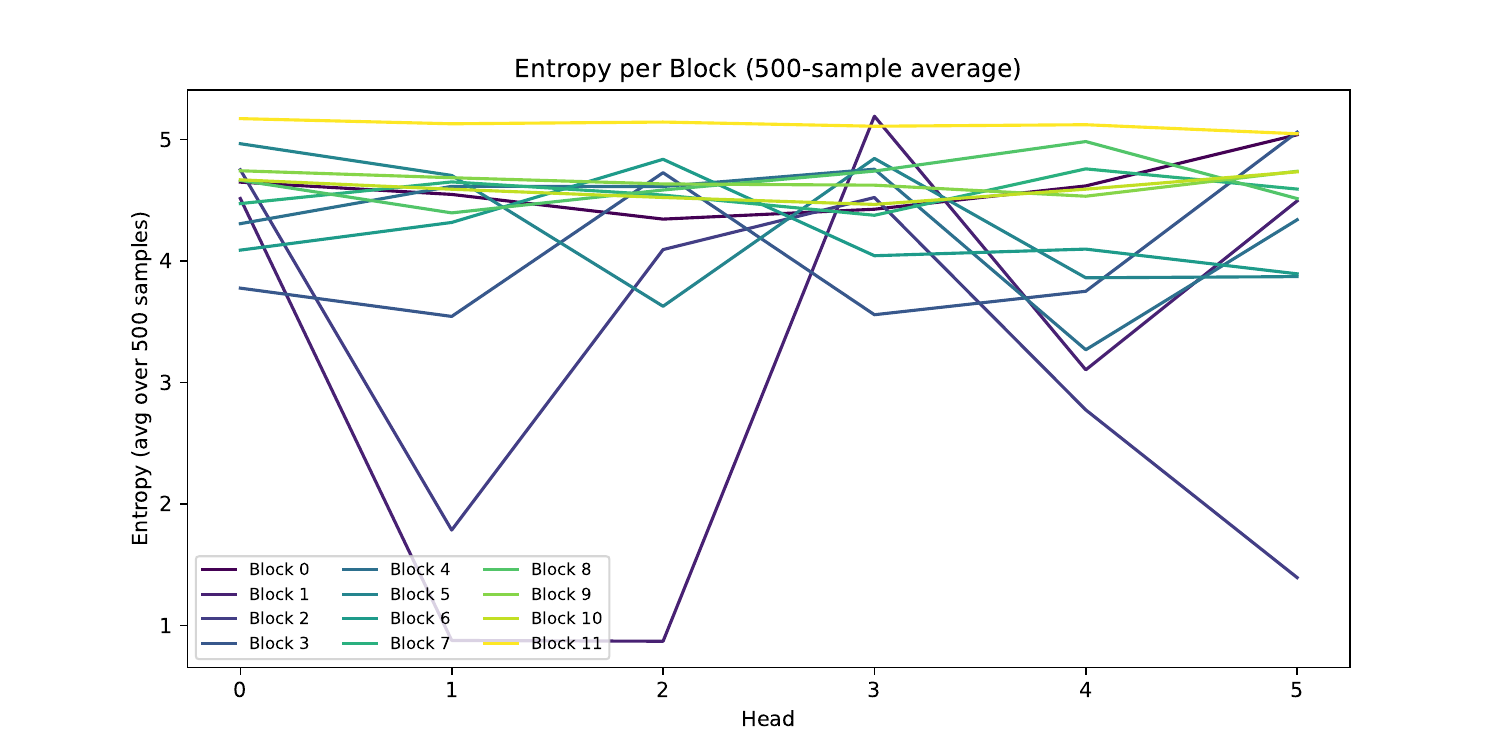}
  \end{minipage}%
  \hfill
  \begin{minipage}{0.5\linewidth}
    \centering
    \includegraphics[width=\linewidth]{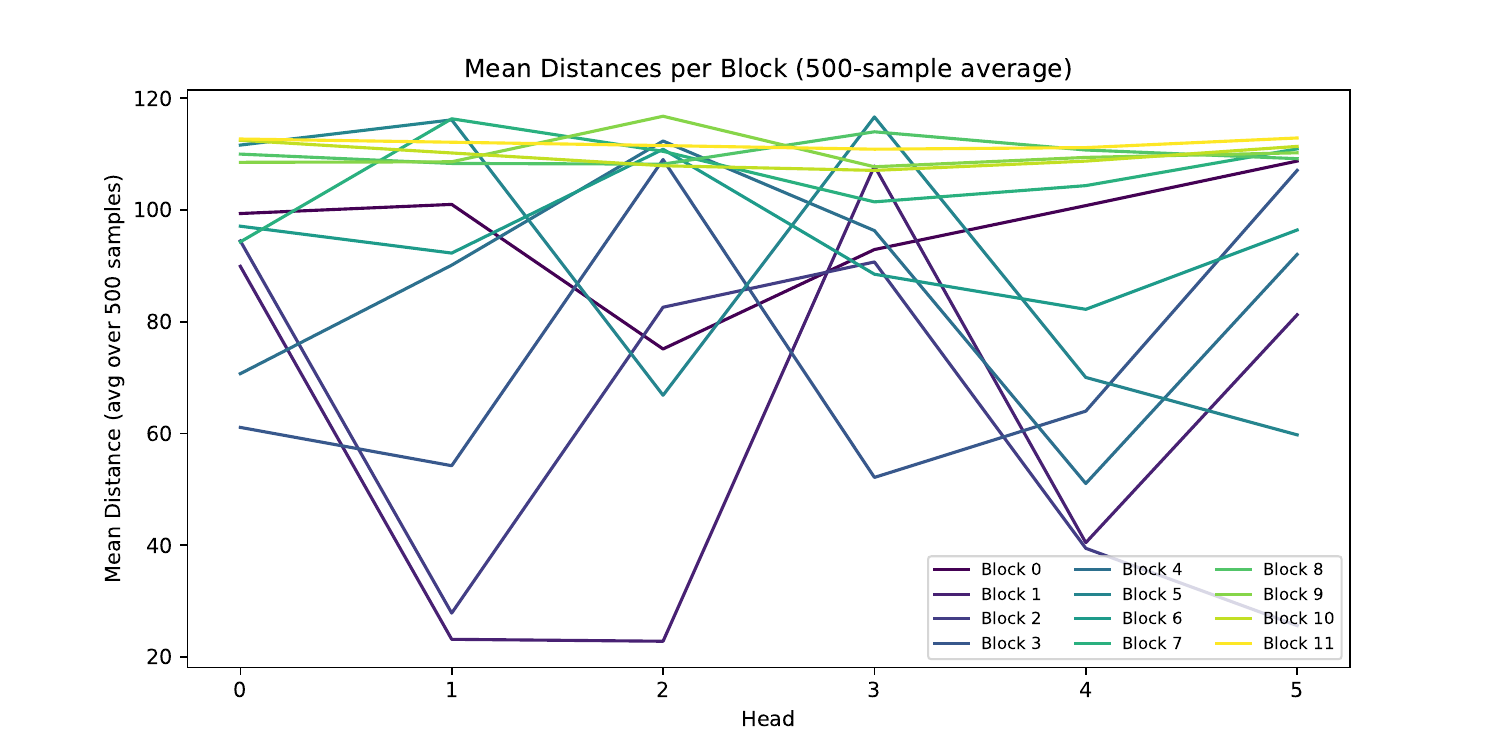}
  \end{minipage}
  \caption{Results for attention entropy (top) and attention distance~\cite{attndist} (bottom). All values represent averages over 500 samples.}
  \label{fig:attndist}
\end{figure}
}
\def\deits{
\begin{table*}[t!]
  \caption{Top-1 accuracy(\%) with DeiT-S on \texttt{ImageNet-100} (\texttt{IN-100}), \texttt{FGVC Aircraft} (\texttt{Aircraft}), and \texttt{Oxford Flowers102}  (\texttt{Flowers}) under different keep rates $r$. The DeiT-S column reports scores without patch pruning. The bold scores indicate the best performance at each keep rate.}
  \label{tab:deits}
  \centering
  \small
  \begin{tabular}{llccccccccc}
  \toprule
  \multicolumn{3}{c}{\textbf{Configurations and baseline}} &
  \multicolumn{7}{c}{\textbf{Keep rate $r$}} \\
  \cmidrule(lr){1-3}\cmidrule(lr){4-10}
  \multicolumn{1}{c}{Dataset} &
  \multicolumn{1}{c}{Method} &
  \multicolumn{1}{c}{DeiT-S} &
  0.9 & 0.8 & 0.7 & 0.6 & 0.5 & 0.4 & 0.3 \\
  \midrule
  \multirow{5}{*}{\texttt{IN-100}}
    & EViT                              & 86.74 & 86.06 & 86.04 & 86.46 & 85.68 & 85.32 & 84.50 & 83.92 \\
    & Ours (Shannon)                & 86.74 & \textbf{86.40} & 85.74 & 86.14 & 85.66 & 85.54 & 84.70 & 84.16 \\
    & Ours (R\'enyi, $\alpha=2.0$)  & 86.74 & 86.34 & \textbf{86.46} & 85.60 & 85.50 & 85.36 & 84.78 & \textbf{84.30} \\
    & Ours (R\'enyi, $\alpha=5.0$)  & 86.74 & 85.88 & 86.20 & 86.30 & \textbf{85.78} & 85.44 & \textbf{85.18} & 84.00 \\
    & Ours (R\'enyi, $\alpha=10.0$) & 86.74 & 86.26 & 86.02 & \textbf{86.72} & 85.72 & \textbf{85.68} & 84.82 & 84.20 \\
  \midrule
  \multirow{5}{*}{\texttt{Aircraft}}
    & EViT                              & 79.72 & 71.29 & 72.10 & 78.88 & 76.30 & 78.82 & 73.69 & 76.96 \\
    & Ours (Shannon)                & 79.72 & 77.53 & \textbf{77.71} & 72.91 & 76.06 & 68.44 & 74.68 & 64.45 \\
    & Ours (R\'enyi, $\alpha=2.0$)  & 79.72 & 71.56 & 67.54 & 77.11 & 71.50 & 73.87 & 80.02 & 70.15 \\
    & Ours (R\'enyi, $\alpha=5.0$)  & 79.72 & 78.94 & 76.75 & \textbf{80.14} & 77.77 & 78.16 & \textbf{80.29} & 74.26 \\
    & Ours (R\'enyi, $\alpha=10.0$) & 79.72 & \textbf{79.84} & 74.32 & 78.55 & \textbf{78.70} & \textbf{80.29} & 77.38 & \textbf{77.08} \\
  \midrule
  \multirow{5}{*}{\texttt{Flowers}}
    & EViT                              & 89.04 & 90.58 & 88.47 & \textbf{90.81} & 87.71 & \textbf{89.67} & \textbf{88.81} & \textbf{86.19} \\
    & Ours (Shannon)                & 89.04 & 90.23 & 84.99 & 88.91 & 87.64 & 86.86 & 86.39 & 84.52 \\
    & Ours (R\'enyi, $\alpha=2.0$)  & 89.04 & 89.64 & 88.26 & 90.63 & 88.49 & 88.76 & 87.04 & 83.61 \\
    & Ours (R\'enyi, $\alpha=5.0$)  & 89.04 & \textbf{90.71} & \textbf{90.14} & 90.45 & \textbf{89.15} & 88.32 & 87.10 & 84.00 \\
    & Ours (R\'enyi, $\alpha=10.0$) & 89.04 & 88.37 & 90.11 & 89.75 & 87.93 & 87.97 & 86.21 & 67.26 \\
  \bottomrule
\end{tabular}

\end{table*}
}

\def\deitb{
\begin{table*}[t!]
  \caption{Top-1 accuracy(\%) with DeiT-B on \texttt{ImageNet-100} (\texttt{IN-100}), \texttt{FGVC Aircraft} (\texttt{Aircraft}), and \texttt{Oxford Flowers102} (\texttt{Flowers}) under different keep rates $r$. The DeiT-B column reports scores without patch pruning. The bold scores indicate the best performance at each keep rate.}
  \label{tab:deitb}
  \centering
  \small
  \begin{tabular}{llccccccccc}
  \toprule
  \multicolumn{3}{c}{\textbf{Configurations and baseline}} &
  \multicolumn{7}{c}{\textbf{Keep rate $r$}} \\
  \cmidrule(lr){1-3}\cmidrule(lr){4-10}
  \multicolumn{1}{c}{Dataset} &
  \multicolumn{1}{c}{Method} &
  \multicolumn{1}{c}{DeiT-B} &
  0.9 & 0.8 & 0.7 & 0.6 & 0.5 & 0.4 & 0.3 \\
  \midrule
  \multirow{5}{*}{\texttt{IN-100}}
    & EViT                              & 86.82 & \textbf{87.04} & \textbf{86.64} & 86.36 & \textbf{86.20} & 85.02 & 84.32 & \textbf{83.82} \\
    & Ours (Shannon)                & 86.82 & 86.48 & 86.58 & 86.18 & 85.38 & 84.80 & 84.12 & 83.02 \\
    & Ours (R\'enyi, $\alpha=2.0$)  & 86.82 & 86.62 & 86.40 & \textbf{86.44} & 85.72 & \textbf{85.20} & 84.40 & 83.14 \\
    & Ours (R\'enyi, $\alpha=5.0$)  & 86.82 & 86.44 & 86.54 & 86.14 & 85.48 & 84.88 & 84.34 & 83.02 \\
    & Ours (R\'enyi, $\alpha=10.0$) & 86.82 & 86.68 & 86.42 & 86.16 & 85.96 & 84.70 & \textbf{84.50} & 83.02 \\
  \midrule
  \multirow{5}{*}{\texttt{Aircraft}}
    & EViT                              & 81.70 & 77.04 & 79.24 & 74.98 & 73.18 & 77.95 & 74.95 & 74.08 \\
    & Ours (Shannon)                & 81.70 & 79.09 & 81.64 & \textbf{83.32} & \textbf{82.24} & 66.85 & 72.55 & 71.68 \\
    & Ours (R\'enyi, $\alpha=2.0$)  & 81.70 & 82.24 & \textbf{83.56} & 71.98 & 67.66 & 71.92 & 66.70 & 76.21 \\
    & Ours (R\'enyi, $\alpha=5.0$)  & 81.70 & \textbf{84.04} & 80.14 & 76.33 & 75.52 & \textbf{83.29} & 72.97 & \textbf{78.31} \\
    & Ours (R\'enyi, $\alpha=10.0$) & 81.70 & 71.44 & 73.99 & 78.52 & 75.04 & 82.09 & \textbf{76.15} & 67.84 \\
  \midrule
  \multirow{5}{*}{\texttt{Flowers}}
    & EViT                              & 93.10 & \textbf{93.36} & 92.58 & 91.88 & 90.94 & 88.01 & 86.49 & 83.72 \\
    & Ours (Shannon)                & 93.10 & 92.75 & 91.59 & \textbf{92.39} & 90.34 & 88.73 & 85.70 & 81.41 \\
    & Ours (R\'enyi, $\alpha=2.0$)  & 93.10 & 92.94 & 92.26 & 91.61 & 90.23 & 88.42 & \textbf{86.99} & 82.92 \\
    & Ours (R\'enyi, $\alpha=5.0$)  & 93.10 & 92.80 & 92.41 & 91.28 & 90.78 & 88.11 & 85.97 & \textbf{84.03} \\
    & Ours (R\'enyi, $\alpha=10.0$) & 93.10 & 93.20 & \textbf{92.83} & 91.74 & 90.36 & \textbf{88.99} & 86.11 & 82.92 \\
  \bottomrule
\end{tabular}

\end{table*}
}

\def\cc{
\begin{table}[t!]
  \caption{Comparison of computational cost with existing methods in terms of the best score of validation top-1 accuracy (acc), FLOPs, and throughput (images/s), by varying the keep rate $r$.}
  \label{tab:cc}
  \centering
  \small
  \begin{tabular}{lccccccc}
    \toprule
     & $r$ & acc(\%) & GFLOPs & images/s \\
    \midrule
    DeiT-S & 1.0 & 86.74 & 4.61 & 1111.4 \\
    \midrule
    \multirow{3}{*}{EViT} & 0.7 & 86.46 \textcolor{blue}{-0.28} & 3.04 \textcolor{blue}{-34.1\%} & 1947.1 \textcolor{blue}{+75.2\%} \\
     & 0.5 & 85.32 \textcolor{blue}{-1.42} & 2.31 \textcolor{blue}{-49.8\%} & 2535.8 \textcolor{blue}{+128.2\%} \\
     & 0.3 & 83.92 \textcolor{blue}{-2.82} & 1.82 \textcolor{blue}{-60.5\%} & 3150.1 \textcolor{blue}{+183.4\%} \\
     \midrule
     \multirow{3}{*}{Ours} & 0.7 & 86.72 \textcolor{blue}{-0.02} & 3.00 \textcolor{blue}{-34.8\%} & 1730.3 \textcolor{blue}{+55.7\%} \\
     & 0.5 & 86.68 \textcolor{blue}{-0.06} & 2.29 \textcolor{blue}{-50.4\%} & 2257.5 \textcolor{blue}{+103.1\%} \\
     & 0.3 & 84.30 \textcolor{blue}{-2.44} & 1.80 \textcolor{blue}{-60.8\%} & 2790.2 \textcolor{blue}{+151.1\%} \\
     \bottomrule
  \end{tabular}
\end{table}
}

\makeatletter
\newcommand{\samethanks}[1][\value{footnote}]{\footnotemark[#1]}
\makeatother
\begin{document}
\title{R\'enyi Attention Entropy for Patch Pruning}
\titlerunning{R\'enyi Attention Entropy for Patch Pruning}
%
\author{Hiroaki Aizawa\thanks{These authors contributed equally to this work.} \and Yuki Igaue\samethanks}
\authorrunning{H. Aizawa and Y. Igaue}
%
\institute{Graduate School of Advanced Science and Engineering\\
Hiroshima University\\
Higashi-Hiroshima, Japan\\
\email{hiroaki-aizawa@hiroshima-u.ac.jp, yuki\_0304@outlook.com}}
\maketitle              
\begin{abstract}
Transformers are strong baselines in both vision and language because self-attention captures long-range dependencies across tokens. However, the cost of self-attention grows quadratically with the number of tokens. Patch pruning mitigates this cost by estimating per-patch importance and removing redundant patches. To identify informative patches for pruning, we introduce a criterion based on the Shannon entropy of the attention distribution. Low-entropy patches, which receive selective and concentrated attention, are kept as important, while high-entropy patches with attention spread across many locations are treated as redundant. We also extend the criterion from Shannon to R\'enyi entropy, which emphasizes sharp attention peaks and supports pruning strategies that adapt to task needs and computational limits. In experiments on fine-grained image recognition, where patch selection is critical, our method reduced computation while preserving accuracy. Moreover, adjusting the pruning policy through the R\'enyi entropy measure yields further gains and improves the trade-off between accuracy and computation.

\end{abstract}
\section{Introduction}
\label{sec:intro}
Deep neural networks, with their strong pattern recognition capabilities, have driven rapid progress across vision, language, and speech. In the field of computer vision, convolutional neural networks leverage local inductive biases from \emph{convolution}, while Transformers~\cite{transformer} learn long-range dependencies directly from data via \emph{self-attention}. These approaches have proved the effectiveness across a wide range of vision tasks.

In particular, the Vision Transformer (ViT)~\cite{vit} is a promising direction and has emerged as a new paradigm. ViT divides an image into fixed-size patches, embeds them as tokens, and uses self-attention to directly handle long-range dependencies among patches. This framework has established ViT-based architecture as strong foundation models for image recognition~\cite{vit,swin,deit}. However, the computational cost of self-attention scales quadratically with the number of patches, hence inference and training costs become expensive for high-resolution inputs and for tasks that require fine-grained discrimination. In spatiotemporal data, similarity between consecutive frames often produces redundant patches. To address these issues, patch pruning, which removes redundant patches that are unnecessary for solving a given task at an early stage in the network, has become key to building practical vision models.

\overview

Estimating which patches to keep remains challenging. Existing pruning criteria mainly rely on summary statistics of attention magnitudes~\cite{evit,spvit,robustpatch,ats}, which do not directly quantify the model’s certainty or the concentration of attention across patches. For example, a patch can receive a high attention score even when attention is broadly dispersed across many locations, which weakens the evidence for retaining that patch. In addition, a large attention weight can coincide with a small contribution from the corresponding value vector~\cite{valuenorm}, in which case the patch should not be prioritized. Consequently, the magnitude-based importance of attention weights does not reliably distinguish uncertainty and can lead to over-retention of redundant patches or erroneous pruning of useful ones.

To resolve these issues, we propose an information-theoretic criterion for patch pruning. Specifically, we use the Shannon entropy~\cite{shannon} of the attention distribution and evaluate patch importance by its entropy. This quantity is often referred to as \emph{attention entropy}. In NLP, attention entropy has been used to improve fairness and to mitigate bias by regularizing or auditing attention patterns~\cite{biasentropy,analyzingbert,eadra}. It is also linked to Transformer training dynamics, where maintaining sufficient attention entropy stabilizes optimization and helps prevent attention collapse~\cite{attncollapse}. When self-attention is selective and concentrated on specific patches, the attention entropy is low. When self-attention is broadly spread across many patches, the attention entropy is high. Furthermore, as shown in Fig.~\ref{fig:overview}, using a DeiT-S~\cite{deit} model pretrained on \texttt{ImageNet-1k}~\cite{imagenet}, we observe that low attention entropy tends to occur on foreground regions, whereas high attention entropy tends to occur on background regions. From this observation, we hypothesize that, especially in image classification, low-entropy patches correspond to object regions and are informative for prediction. We therefore design a pruning policy that preferentially keeps low-entropy patches and treats high-entropy patches, whose attention is dispersed across many locations, as candidates for removal. 

Moreover, prior work suggests that attention entropy shapes how language models balance generality and specificity~\cite{analyzingmhsa,analyzingbert}. This motivates a criterion with tunable sensitivity to concentration, and we adopt R\'enyi entropy. R\'enyi entropy~\cite{renyi} is a generalization of Shannon entropy with an order parameter that controls peak emphasis. By adjusting this order, the score can accentuate sharp peaks and tune pruning strength to task characteristics. These entropy-based criteria require only a single forward pass, are easy to implement, need no additional training, and integrate cleanly with existing ViTs. We refer to our method as \emph{R\'enyi Attention Entropy Pruning}.

In the experiments, we validated the effectiveness on both standard and fine-grained image classification. Especially, using a pre-trained DeiT-S and evaluating on \texttt{ImageNet-100}~\cite{imagenet100}, our approach reduces computation by about 35\% and improves inference speed by about 55\% while limiting the accuracy drop to about 0.02 percentage points relative to no pruning. Under comparable computational cost, it achieved higher accuracy than prior patch pruning methods such as EViT~\cite{evit}. In addition, adapting the pruning policy through the peak emphasis of R\'enyi entropy further improves the trade-off between accuracy and computation. In summary, our main contributions are as follows:
\begin{itemize}
  \item \textbf{Information-theoretic patch pruning.} We introduce information-theoretic criteria, Shannon and R\'enyi attention entropies for patch pruning.
  \item \textbf{R\'enyi attention entropy.} Our R\'enyi attention entropy enables flexible control over peak emphasis and pruning strength, supporting task- and budget-aware policies.
  \item \textbf{Task-oriented pruning.} Comprehensive evaluations on \texttt{ImageNet-100}, \texttt{FGVC Aircraft}, and \texttt{Oxford Flowers102} demonstrated favorable accuracy-computation trade-offs and consistent gains from entropy-driven pruning.
\end{itemize}

\section{Related Work}

\subsection{Patch Pruning for Efficient ViTs}
Patch pruning has been widely studied to reduce the computation of self-attention for efficient Vision Transformers. Many approaches follow a pipeline that uses a pretrained model, estimates per-patch importance, and retains informative patches while pruning redundant ones. Among these stages, estimating patch importance plays a central role. Scores have been derived from learnable subnetworks with trainable parameters~\cite{dynamicvit}, summary statistics of attention weights~\cite{evit,spvit,ats,evovit}, head-wise variance~\cite{robustpatch}, predicted probabilities~\cite{avit}, and inter-patch similarity~\cite{intrapatch}. Rather than discarding information from redundant patches, some studies merge them to preserve content while reducing the token count~\cite{evit,tome,spvit,evovit}. Task-aware pruning tailored to specific applications such as segmentation has also been explored~\cite{dtop,cropr}. Although these methods achieve strong computational efficiency while maintaining performance, they often overlook properties of the attention distribution such as model confidence and dispersion. In this work, we quantify the attention distribution with an information-theoretic criterion, attention entropy, and use it to estimate patch importance for patch pruning.

\proposed

\subsection{An Information-Theoretic View of Transformers}
Information theory, established by Shannon, treats entropy as a fundamental measure of uncertainty in probability distributions~\cite{shannon}. Applied to Transformer self-attention, this defines an attention entropy for each query that is low when the distribution is concentrated and high when it is diffuse. In natural language processing, attention entropy has been used to improve interpretability~\cite{attnkey,analyzingbert}, mitigate bias~\cite{biasentropy}, and enhance neural machine translation~\cite{eadra} by regularizing attention patterns. Moreover, several studies suggest that attention entropy is related to how language models balance generality and specificity~\cite{analyzingmhsa,analyzingbert}. It has also served as a lens on training dynamics, where entropy behaves consistently with model confidence and excessively low entropy has been reported to induce instability and attention collapse~\cite{attncollapse}. In vision, work that leverages attention entropy has progressed, especially in semantic segmentation~\cite{attnsemseg}, yet its direct use for scoring patch importance remains underexplored. We focus on attention entropy as an importance estimator for patch pruning, show that it is a meaningful criterion in visual tasks, and point to new directions for leveraging attention entropy in vision.

\section{Methodology}
\label{sec:methodogy}
In this section, we formulate our proposed attention entropy-based patch pruning strategy. First, we review the Vision Transformer model~\cite{vit} in Section~\ref{ssec:transformer} and then define Shannon and R\'enyi attention entropies in Section~\ref{ssec:attn_ent}. Finally, we present our patch pruning framework utilizing these attention entropies as patch importance in Section~\ref{ssec:patch_pruning}.

\subsection{Vision Transformer}
\label{ssec:transformer}
We formulate the Vision Transformer (ViT)~\cite{vit} and its multi-head self-attention (MHSA). Given an input image, let the set of patch tokens be $\{\boldsymbol{x}_i \in \mathbb{R}^d\}_{i=1}^{n-1}$ and let $\boldsymbol{x}_{\texttt{class}}\in\mathbb{R}^d$ denote the $\texttt{class}$ token. We collect them into 
\begin{align}
\boldsymbol{X}=\big[\boldsymbol{x}_{\texttt{class}},\, \boldsymbol{x}_{1},\dots,\boldsymbol{x}_{n-1}\big]\in \mathbb{R}^{d\times n},
\end{align}
This matrix is processed block by block, where each block consists of MHSA, a feed-forward network, and layer normalization~\cite{layernorm}. The blocks model contextual relationships across the tokens corresponding to image patches.

MHSA carries out this modeling. Concretely, for a single attention head, using learnable Query and Key matrices $\boldsymbol{W}_Q, \boldsymbol{W}_K \in \mathbb{R}^{d\times d'}$, we compute the attention weights
\begin{align}
\mathrm{Attn}(\boldsymbol{X})
:=\sigma\!\left(\frac{\boldsymbol{X}^{\top}\boldsymbol{W}_K \boldsymbol{W}_Q^{\top}\boldsymbol{X}}{\sqrt{d'}}\right)\in\mathbb{R}^{n\times n},
\label{eq:attn}
\end{align}
where $\sigma(\cdot)$ denotes the softmax. Therefore, each row encodes the relation of a patch token $\boldsymbol{x}_i$ to all other tokens. In ViT, from this attention matrix and the token matrix obtained via a learnable Value matrix $\boldsymbol{W}_V\in\mathbb{R}^{d\times d'}$, the attended token matrix $\boldsymbol{X}'$ is extracted as
\begin{align}
\boldsymbol{X}'=\boldsymbol{W}_V^{\top}\boldsymbol{X}\,\mathrm{Attn}(\boldsymbol{X}) \in \mathbb{R}^{d'\times n}.
\end{align}

In MHSA with $h$ heads, we compute the above per head and then concatenate the head outputs followed by an output projection $\boldsymbol{W}_O\in\mathbb{R}^{d \times d'}$. With a residual connection, the token matrix propagated to the next block is
\begin{align}
\bar{\boldsymbol{X}}=\boldsymbol{X}+\big[\mathrm{Concat}_{\text{heads}}(\boldsymbol{W}_O\boldsymbol{X}')\big].
\end{align}
For simplicity, we omit layer normalization and the feed-forward sublayer in the equations. After $L$ such blocks, the classification layer uses the final class token to produce class probabilities through a linear layer and a softmax function.

\subsection{Attention Entropy as Patch Importance}
\label{ssec:attn_ent}

We begin by formalizing the per-patch attention distribution that arises from the softmax in the attention computation (Eq.~\eqref{eq:attn}) and then define attention entropies based on Shannon entropy~\cite{shannon} and R\'enyi entropy~\cite{renyi}.

First, the attention distribution is given by the following definition.

\begin{definition}[Patch attention distribution]\label{def:attn_dist}
Let $\boldsymbol{X}_{patches}=\big[\boldsymbol{x}_1,\dots,\boldsymbol{x}_{n-1}\big]\in\mathbb{R}^{d \times (n-1)}$ be the matrix of patch tokens and $\boldsymbol{W}_Q,\boldsymbol{W}_K\in\mathbb{R}^{d\times d'}$ be learnable projection matrices. For the $i$-th patch $\boldsymbol{x}_i$, its attention distribution over patch tokens $\boldsymbol{a}(\boldsymbol{x}_i)\in\mathbb{R}^{n-1}$ is
\begin{align}
\boldsymbol{a}(\boldsymbol{x}_i)
:= \sigma\!\left(\frac{\boldsymbol{X}_{patches}^{\top}\boldsymbol{W}_K \boldsymbol{W}_Q^{\top} \boldsymbol{x}_i}{\sqrt{d'}} \right),
\end{align}
where $\sigma(\cdot)$ is the softmax function and $\boldsymbol{a}(\boldsymbol{x}_i)=\big[a_{1}(\boldsymbol{x}_i),\ldots,a_{n-1}(\boldsymbol{x}_i)\big]^\top$ with $\sum_{j=1}^{n-1} a_j(\boldsymbol{x}_i)=1$.
\end{definition}

In practice this distribution is computed per head. Unless otherwise noted we aggregate head-wise quantities by a simple average for robustness across heads.

\visattnent

Given Definition~\ref{def:attn_dist}, we quantify the uncertainty of attention based on Shannon entropy~\cite{shannon} as follows.

\begin{definition}[Shannon attention entropy]
The Shannon attention entropy of the $i$-th patch is
\begin{align}
\mathcal{H}(\boldsymbol{x}_i)
:= - \sum_{j=1}^{n-1} a_{j}(\boldsymbol{x}_i)\,\log a_{j}(\boldsymbol{x}_i),
\end{align}
where $\log$ is the natural logarithm.
\end{definition}

Shannon attention entropy attains its maximum for the uniform distribution and its minimum $0$ when all mass concentrates on a single token. Hence higher values indicate attention dispersed over many tokens and lower values indicate attention concentrated on specific patches. 

To control peak emphasis in a principled way we also consider a R\'enyi attention entropy, which is a generalization of Shannon attention entropy.

\begin{definition}[R\'enyi attention entropy]\label{def:renyi_attn_ent}
For order $\alpha>0$ with $\alpha\neq 1$, the R\'enyi attention entropy of the $i$-th patch is
\begin{align}
\mathcal{H}_{\alpha}(\boldsymbol{x}_i)
:= \frac{1}{1-\alpha}\,\log \sum_{j=1}^{n-1} a_{j}(\boldsymbol{x}_i)^{\alpha}.
\end{align}
\end{definition}

As the order approaches one, $\mathcal{H}_{\alpha}(\boldsymbol{x}_i)$ converges to the Shannon attention entropy. When $0<\alpha<1$ lower-probability events receive relatively larger weight and dispersion is emphasized. When $\alpha>1$ higher-probability events are emphasized and concentration is accentuated. 

As a preliminary experiment, we visualize the attention entropy of each patch using a DeiT-S~\cite{deit} model pretrained on \texttt{ImageNet-1k}~\cite{imagenet}. As shown in Fig.~\ref{fig:visattnent}, the visualizations reveal a clear trend: attention entropy is high in background regions and low in object regions. Therefore, we treat low-entropy patches as informative and retain them, while high-entropy patches are regarded as redundant candidates for pruning.

\deits

\subsection{R\'enyi Attention Entropy Pruning}
\label{ssec:patch_pruning}

Building on these observations, we propose patch pruning based on Shannon and R\'enyi attention entropies, as shown in Fig.~\ref{fig:proposed}. We refer to our method as \emph{R\'enyi Attention Entropy Pruning}. Following prior approaches~\cite{evit,tome}, our method applies the following patch pruning during the fine-tuning phase on the target dataset after building a pretrained model.

\begin{enumerate}
    \item \textbf{Computing attention distribution and entropies.} The input sequence $\boldsymbol{X}$ is processed in each Transformer block. The softmax yields an attention distribution $\boldsymbol{a}(\boldsymbol{x}_i)$ for each patch and we quantify it using Shannon or R\'enyi attention entropy. In practice, we compute it per head and average across heads to obtain a per-layer score.
    \item \textbf{Pruning policy based on importance.} Lower entropy indicates higher importance and such patches are retained, while higher entropy patches are treated as redundant candidates for pruning. We set a fixed keep ratio $k$ before fine-tuning. Control tokens such as \texttt{class} are always retained. When using R\'enyi entropy, the peak emphasis is adjusted to match task characteristics and the compute budget.
    \item \textbf{Layerwise pruning and propagation.} Only the selected tokens are propagated to the next layer, and the same procedure is repeated. Finally, prediction is made from the \texttt{class} token, achieving reduced computation while preserving accuracy.
\end{enumerate}

\section{Evaluation}
\subsection{Evaluation Protocol}
We evaluate patch pruning based on Shannon and R\'enyi attention entropy on both generic and fine-grained image classification. For generic classification, we use the standard \texttt{ImageNet-100}~\cite{imagenet100} benchmark, and for fine-grained classification we adopt \texttt{FGVC Aircraft}~\cite{aircraft} and \texttt{Oxford Flowers102}~\cite{flowers}. Across these tasks, our attention entropy criterion identifies informative patches more effectively than the attention weight-based criterion used in EViT~\cite{evit}. We also show that the balance between generality and specificity in attention, as reflected by attention entropy~\cite{analyzingmhsa,analyzingbert}, can be tuned using R\'enyi attention entropy, with the clearest gains observed on the fine-grained benchmarks.

\subsubsection{General Settings.} Following the patch pruning-based training protocol~\cite{evit}, we built our method on a DeiT-S model~\cite{deit} pretrained on \texttt{ImageNet-1k}~\cite{imagenet} and fine-tuned all models for 100 epochs using AdamW~($\beta_1 = 0.9$, $\beta_2 = 0.999$, weight decay$=0.05$)~\cite{adamw}. Fine-tuning was conducted with a batch size of 256 on a single NVIDIA RTX A6000 GPU. During training and inference, input images were resized to a resolution of $224 \times 224$. For data augmentation during training, we applied RandomCrop, RandomHorizontalFlip, Mixup~\cite{mixup}, CutMix~\cite{cutmix}, and RandomErasing~\cite{erasing}. Furthermore, we measured inference throughput and FLOPs using \texttt{fvcore}.

\subsubsection{Pruning Settings.} Following the pruning protocol~\cite{evit}, we prune redundant patches only at blocks $\{4,7,10\}$ during fine-tuning and at test time. For our Shannon and R\'enyi attention entropies, we search orders $\alpha=\{2.0, 5.0, 10.0\}$. Shannon entropy corresponds to order $\alpha=1.0$. In each block, we sort patch tokens by their scores and keep the top $r$ fraction of tokens, where $r \in \{0.9, 0.8, ..., 0.3\}$. We refer to $r$ as the keep rate. The \texttt{class} token is always kept.

\deitb

\vispatch
\attnhist

\subsection{Main Results}
\subsubsection{Results on \texttt{ImageNet-100}.} We discuss the proposed attention entropy patch pruning using DeiT-S trained on \texttt{ImageNet-100}, which requires general image discrimination. Following the experimental protocol described above, we compared our method with EViT~\cite{evit}. The main results are in Table~\ref{tab:deits}. For each keep rate $r$, we selected the best among Shannon and R\'enyi and observed consistent gains over EViT. As representative cases, our method improved top-1 accuracy by $+0.42$ at $r=0.8$ and by $+0.38$ at $r=0.3$. At $r=0.7$ the accuracy reached 86.72, which was within 0.02 of the no-pruning baseline 86.74 while reducing the number of tokens by 30\%. We further discuss the choice of entropy within our method. Shannon performed best at mild pruning such as $r=0.9$. R\'enyi with a larger order tended to be preferable at moderate pruning such as $r=0.7$ and $r=0.5$. These trends supported our design that peak emphasis in R\'enyi entropy could be tuned to the token budget. Overall, attention entropy provided a more effective signal than the attention-magnitude criterion in EViT.

\subsubsection{Results on \texttt{FGVC Aircraft}.} Next, we compared our approach with EViT using DeiT-S trained on \texttt{FGVC Aircraft}, which requires specific and fine-grained discrimination. From Table~\ref{tab:deits}, our method consistently achieved superior or comparative performance to EViT at several keep rates. As a representative case, at $r=0.4$, R\'enyi with $\alpha=5.0$ achieved 80.29, which was higher than both EViT at the same keep rate and the baseline 79.72 while using only 40\% of the tokens. At $r=0.7$ the accuracy reached 80.14, again higher than EViT and above the baseline. We observed that larger R\'enyi orders were generally preferable on this dataset, which aligns with the need to emphasize sharp, part-level cues for specificity.

\cc

\subsubsection{Results on \texttt{Oxford Flowers102}.}
For the other fine-grained dataset, \texttt{Oxford Flowers102}, we also summarized the results in Table~\ref{tab:deits}. The results show that our method was competitive with EViT under mild to moderate pruning. At $r=0.9$ R\'enyi with $\alpha=5.0$ reached 90.71, slightly above EViT, and at $r=0.8$ it reached 90.14, clearly above EViT. At $r=0.7$ our best result 90.63 was close to EViT. Under stronger pruning such as $r \le 0.5$ EViT tended to be superior on this dataset. These results indicate that tuning the R\'enyi order is important for adapting the attention entropy criterion to dataset characteristics.

\subsubsection{Summary.}
Across both general and fine-grained tasks, we found that the accuracy varied considerably with the choice of $\alpha$ even at the same keep rate $r$. This parameter $\alpha$ should be tuned for each keep rate, and the resulting gains justify the tuning overhead.

\subsection{Ablation Study}
\subsubsection{Scaling to Larger Models.} 
As an ablation study, we report results on DeiT-B in Table~\ref{tab:deitb}. Except for specific settings such as $r=\{0.3,0.6,0.8,0.9\}$ on \texttt{ImageNet-100} and $r=0.9$ on \texttt{Oxford Flowers102}, the proposed method achieved comparable or superior performance to existing approaches, consistent with the results observed on DeiT-S. In DeiT-B, in particular, we observed notable improvements on fine-grained datasets, \texttt{FGVC Aircraft}, further demonstrating the applicability of the proposed method.

\subsubsection{Visualization of Informative Patches.}
For better understanding of the proposed method, we qualitatively analyze the behavior of the proposed entropy criterion by visualizing informative and pruned patches. Figure~\ref{fig:vispatch} shows patch pruning results at the keep rate $r=0.7$ for EViT and our method. EViT often captures foreground regions that are useful for classification, yet it also shows scattered patch selections in background areas and removal of foreground patches at early blocks. In contrast, our R\'enyi attention entropies consistently focus on foreground regions and captures the part-dependent cues needed for fine-grained recognition. Since the R\'enyi order controls the peak emphasis of the attention distribution, it allows the selection bias to be adjusted to the task.

\attndist

\subsubsection{Analysis of Computational Cost.}
The trade-off between accuracy and computational cost is one of the critical factors, particularly in practical applications. We conducted experiments using DeiT-S as the base model and compared the results against the case without pruning, as well as between our proposed method and EViT, in terms of the best score of validation top-1 accuracy (acc), FLOPs, and throughput (images/s), varying the keep rate $r=\{0.7,0.5,0.3\}$. For our method, we reported the values obtained under the condition $\alpha$ that achieved the highest accuracy for each keep rate. The results are summarized in Table~\ref{tab:cc}. Across all keep rates, our method consistently outperformed EViT by achieving higher accuracy with fewer FLOPs. In particular, at $r=0.7$, throughput improved by approximately 55\% while the accuracy drop was limited to about 0.02\%. On the other hand, in terms of throughput, our method showed slightly lower performance than EViT.

\subsubsection{Analysis of R\'enyi Entropy Distribution.}
We deeply analyze the attention entropy distribution for patch pruning, as shown in Fig.~\ref{fig:attnhist}. The histograms reveal that increasing the R\'enyi order $\alpha$ shifts the overall entropy distribution toward lower values and changes its shape across blocks. This indicates that the order controls the selection bias by re-ranking tokens under the same keep rate for better pruning. In addition, our new finding is that the attention entropy distribution changes with depth. In particular, later blocks are skewed toward specific entropy ranges. This observation is further supported by the entropy heatmap visualization for each block (see Fig.~\ref{fig:visattnent}).

\subsubsection{Relationship between Attention Entropy and Attention Distance.}
We analyze attention entropy from the perspective of feature extraction. Fig.~\ref{fig:attndist} compares attention entropy with attention distance~\cite{attndist}, which represents the receptive field size of ViTs. As is evident from the figure, these metrics show clear correlations with respect to head diversity and block depth. While a deeper analysis of these relationships is left for future work, these results suggest that attention entropy can serve not only as a criterion for patch pruning but also as a regularizer for training vision models. They also motivate adaptive tuning that aligns the keep rate $r$ and the R\'enyi entropy order $\alpha$ with task characteristics, enabling task-aware pruning without additional retraining, applicable both during training and at inference.

\section{Conclusion}
We introduced an information-theoretic criterion for patch pruning that identifies informative tokens using the Shannon attention entropy of their attention distributions. The score is computed in a single forward pass, requires no additional training, and integrates straightforwardly with existing ViTs. On general and fine-grained tasks, our method achieved accuracy that was superior or competitive with EViT across a wide range of keep rates. Our analyses showed that low attention entropy aligns with foreground regions while high attention entropy aligns with background, which supports using attention entropy as an importance signal. We also found that the attention entropy distribution varies with depth. This observation motivates dynamically adjusting the R\'enyi order and the keep rate per block and per instance in future work. We believe that this information-theoretic view of attention entropy opens up new directions for a broad range of vision tasks.  

\bibliographystyle{splncs04}
\bibliography{main}
\end{document}